\documentclass[letterpaper, 10 pt, conference]{ieeeconf}  %

\IEEEoverridecommandlockouts                              %

\usepackage{amsmath,amssymb}
\usepackage{booktabs}
\usepackage{tabularx}
\usepackage{graphicx}
\usepackage{xcolor}
\makeatletter\let\NAT@parse\undefined\makeatother
\usepackage{hyperref}
\usepackage{cite}
\usepackage{float}
\usepackage{balance}
\usepackage{fontawesome5}

\newcommand{\Xz}{X^0} %
\newcommand{\Xo}{X^1} %
\newcommand{\Asym}{\mathcal{A}_{\text{sym}}}
\newcommand{\Cclass}{\mathcal{C}}

\title{\LARGE \bf
Flow Policies as Actions of Skill-Level World Models: Learned and Symbolic Abstractions for Long-Horizon Planning
}

\author{Andreu Matoses Gimenez$^{*,1}$, Andrei-Carlo Papuc$^{*,1}$, Chris Pek$^{1}$, and Javier Alonso-Mora$^{1}$%
\thanks{This work was supported by the European Union through the ERC INTERACT, Grant 101041863, and by the Office of Naval Research Global under Grant N62909-25-12027 (Project SECURE).
        Views and opinions expressed are, however, those of the author(s) only and do not necessarily reflect those of the European Union.
        Neither the European Union nor the granting authority can be held responsible for them.}%
\thanks{*Equal contribution.}%
\thanks{$^{1}$The authors are with the Cognitive Robotics Department, Delft University of Technology, Delft, The Netherlands
        {\tt\small \{A.MatosesGimenez, A.C.Papuc, C.Pek, J.AlonsoMora\}@tudelft.nl}}%
\thanks{This work has been submitted to the IEEE for possible publication. Copyright may be transferred without notice, after which this version may no longer be accessible.}%
}

\begin{document}

\maketitle
\thispagestyle{empty}
\pagestyle{empty}

\begin{abstract}

Latent world models enable robots to plan by predicting the consequences of actions.
Planning long tasks with control-rate actions requires many prediction steps, which enlarges the search space and accumulates error.
Skill-level actions shorten these sequences, but a symbolic skill vocabulary requires domain knowledge and labeled demonstrations.
We construct skill-level actions from the inputs of a flow-matching policy trained on demonstrations segmented into complete skills.
The policy maps a noise seed and an observation, optionally with a code or label, to a complete skill execution, so one execution is one world-model transition.
On this mechanism we propose four action abstractions with increasing task knowledge: a compressed seed, two discrete codes learned from the demonstrations, and a symbolic label.
We evaluate them with a common world-model training procedure and planning framework on simulated block rearrangement tasks that require up to 14 sequential skills.
The symbolic label succeeds in over 90\% of the tasks that require up to six skills and degrades beyond.
Without any label, an object-centric learned code matches it on single-skill tasks and retains half to three quarters of its success on tasks of two to five skills.
Ablations attribute much of the label's advantage to its planner knowing which actions are applicable, rather than to the label itself.
Beyond six skills the search, not the world model, limits success.
Project page:
\faGlobe~{\urlstyle{same}\href{https://andreumatoses.github.io/research/flow-skill-wm}{\nolinkurl{andreumatoses.github.io/research/flow-skill-wm}}}

\end{abstract}

\section{Introduction}
\label{sec:intro}

Long-horizon manipulation tasks, such as rearranging block towers, require sequences of skills whose outcomes determine which actions can follow.
Planning directly over motor commands involves many prediction steps and a large search space.
Task-and-motion planning addresses this difficulty by combining high-level actions, such as moving or stacking a block, with motion planning \cite{garrett2021tamp}.
These high-level actions, called symbolic operators, describe what an action does and when it can be performed.
For example, moving a block requires that no other block rests on it.
Specifying these actions and rules requires domain knowledge.

Latent world models offer a learned way to predict the consequences of candidate actions.
They allow a planner to evaluate action sequences in a compact, learned representation of the scene before executing them \cite{lecun2022jepa,assran2023ijepa}.
The choice of action abstraction determines the temporal scale of these predictions: motor commands describe individual control steps, whereas skills describe complete behaviors.
A central question is how to construct skill-level actions that a world model can predict and a robot can execute.

\begin{figure}[!t]
  \centering
  \includegraphics[width=\linewidth]{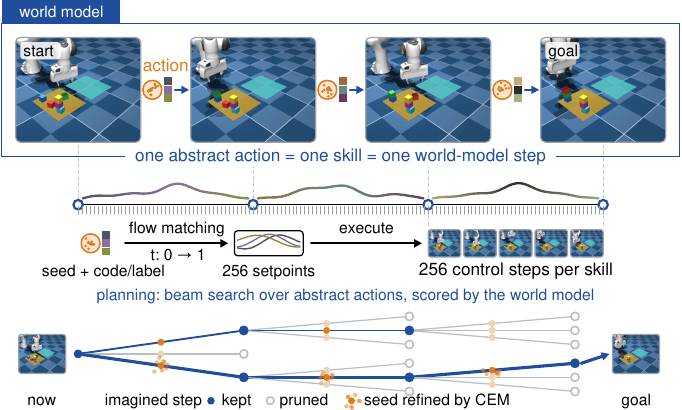}
  \caption{
    \textbf{Planning over complete skills.} The world model predicts the outcome of an entire skill in one step.
    Each planned action selects inputs to a flow policy, which generates 256 control steps.
    The planner evaluates candidate skill sequences using the world model, executes the first selected skill, and replans from the observed scene.
  }
  \label{fig:page1-teaser}
\end{figure}

We investigate this question using flow-matching policies trained on demonstrations of complete skills \cite{lipman2023flowmatching}.
A flow policy generates a sequence of motor commands from the current observation and a sampled noise input, called a \emph{seed}.
It can also receive a code or label that specifies the intended behavior.
The planner selects these inputs, while the policy generates the motion.
We treat each complete policy execution as one world-model transition between settled scenes, where the robot and objects are at rest (Fig.~\ref{fig:page1-teaser}).

Within this framework, we compare four action abstractions: a compressed seed of the policy; a discrete code learned from the scenes before and after a demonstrated skill; an object-centric learned code describing which object moves and where it goes; and a hand-written symbolic label specifying the skill, object, and target.
Each abstraction provides inputs from which the policy generates the robot's motion.
All four use demonstrations divided into complete skills, but the first three do not require demonstrations to be annotated with symbolic action labels.

Planning also requires deciding which actions to consider at each step.
The abstractions differ in how these candidates are selected.
The symbolic label uses hand-written rules, called \emph{preconditions}, to determine which actions can be performed, and \emph{effects} to describe how they change the scene.
The learned-code abstractions instead use learned models to estimate which actions can be performed and their effects.
We therefore ask how both the action abstraction and the selection of candidate actions affect planning as the required number of skills increases.
We study these factors in a simulated block rearrangement problem using a shared policy architecture, world-model training procedure, and planning framework.

Our contributions are:
\begin{itemize}
  \item A skill-level world-modeling framework in which each
  transition corresponds to a complete flow-policy execution
  (\S\ref{sec:flow-actions}).

  \item Four action abstractions spanning compressed seeds,
  learned codes, and symbolic labels, evaluated on block
  rearrangement tasks of increasing difficulty
  (\S\ref{sec:wm}, \S\ref{sec:results}).

  \item Ablations and diagnostics examining how candidate action
  selection, skill execution, and world-model prediction affect
  planning success (\S\ref{sec:results}).
\end{itemize}

\begin{figure*}[!t]
  \centering
  \includegraphics[width=\linewidth]{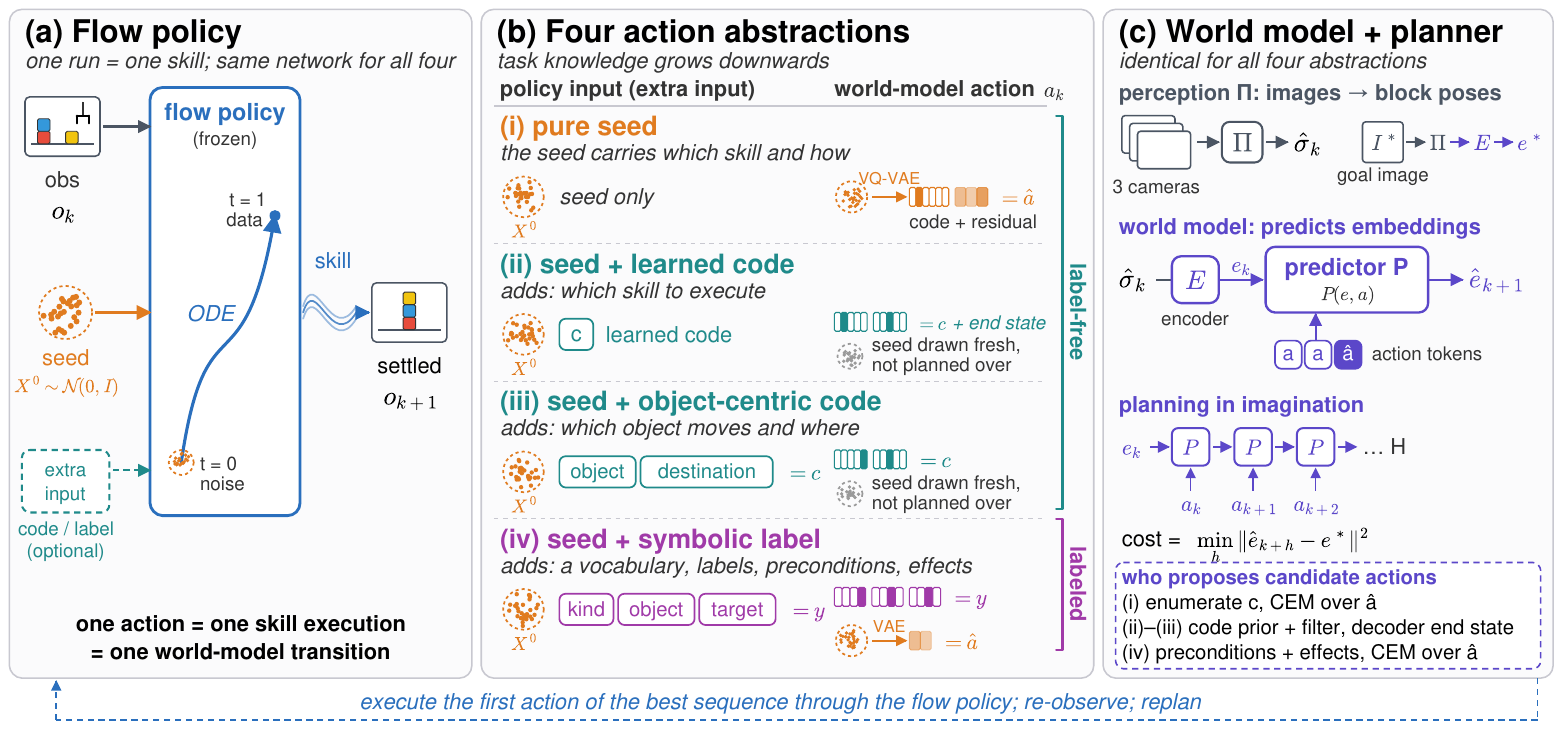}
  \caption{
    \textbf{Skill-level world modeling and planning.} (a) A flow policy generates a complete skill from the current observation, a noise seed, and an optional code or label.
    Each abstraction uses a separately trained policy with a common architecture.
    (b) The four action abstractions: a compressed seed, a learned code, an object-centric code, and a symbolic label with a continuous residual.
    The first three require no symbolic action labels.
    For the two learned-code abstractions, the seed is sampled at execution rather than selected by the planner.
    (c) A separately trained world model predicts the outcomes of candidate skill sequences in a learned representation of the scene.
    Candidates are proposed by enumeration, by learned models, or by symbolic rules, depending on the abstraction.
    The planner scores sequences by their predicted distance to the goal, executes the first selected skill, observes the resulting scene, and replans.
  }
  \label{fig:teaser}
\end{figure*}

\section{Related Work}
\label{sec:related}

\smallskip\noindent\textbf{World models and temporal abstraction.} Latent world models support control by predicting the consequences of candidate actions in a learned representation of the scene \cite{zhou2025dinowm,assran2025vjepa2}.
TD-MPC2 combines short model rollouts with a learned value estimate of the final rollout state \cite{hansen2024tdmpc2}.
Others extend the horizon by predicting the effects of longer actions: CAVIN separates high-level effects from low-level motion generation \cite{fang2019dynamics}, and HWM learns world models at multiple temporal scales, compressing sequences of primitive actions into latent actions and using predicted subgoals to guide lower-level planning \cite{zhang2026hwm}.
We instead set the prediction timescale through a flow policy trained on demonstrations of complete skills: one policy execution is one world-model transition, and the policy itself produces the underlying motor commands.
Planning thus operates over skills rather than control steps, reducing the number of predictions required for long-horizon tasks.

\smallskip\noindent\textbf{Generative policies and planning.} Diffusion and flow policies can represent multiple behaviors for the same observation \cite{chi2023diffusionpolicy,lipman2023flowmatching,black2024pi0}, and Diffuser uses a generative trajectory model directly for planning \cite{janner2022diffuser}.
DSRL shows that the input noise of a frozen diffusion or flow policy controls its behavior, and uses reinforcement learning to select it \cite{wagenmaker2025steering}.
DiWA fine-tunes diffusion policies with reinforcement learning on experience imagined by a learned world model \cite{chandra2025diwa}.
We likewise keep the policy fixed, but search over its inputs with a separate world model at planning time.
These inputs form an executable action abstraction: the planner selects a seed and, where applicable, a code or label, and the policy turns them into a complete skill.
The challenge is to represent these inputs in a form that supports both reliable execution and prediction of their effects.

\smallskip\noindent\textbf{Learned and symbolic action abstractions.}
Genie, LAPO, and LAPA infer latent actions from changes between observations without action annotations \cite{bruce2024genie,schmidt2024lapo,ye2025lapa}.
Our learned codes also capture changes between observations, but a flow policy trained on demonstrated motor commands executes the corresponding skills.
This removes the need for symbolic action labels, though demonstrations must still be segmented into complete skills.
Task-and-motion planning uses symbolic operators that describe actions and their preconditions \cite{garrett2021tamp}; such symbols can also be learned from given skills \cite{konidaris2018symbols}.

Building on these ideas, we seek an action abstraction that lets a world model predict the effects of complete skills and a robot execute them.
We study four such abstractions, combining learned codes and symbolic structure, to identify what supports long-horizon planning and what continues to limit it.

\section{Action Abstractions over a Flow Policy}
\label{sec:flow-actions}

A flow-matching policy maps a noise sample, the \emph{seed}, and an observation to a complete skill execution.
One execution is one world-model transition, and the action is what the planner chooses among the policy's inputs, from a compressed seed to a symbolic label.
This section defines the policy, the four abstractions, and the properties they must satisfy.

\subsection{The flow policy and its seed}
\label{sec:prelim}

Demonstrations are segmented into skills, each recorded as an \emph{action chunk} $\Xo \in \mathbb{R}^{T \times d}$, a sequence of $T$ low-level commands with $d$ values each, paired with the observation $o$ at its start; superscript $1$ marks data and $0$ noise, the flow-matching convention.
The observation contains the poses of all objects and the robot state, and \emph{no goal, task, or label of any kind}.
Because $o$ carries no signal about which skill follows, the demonstrations from one start state must include several valid skills; otherwise the policy interpolates between them and commits to none.
Segmentation into skills is the one assumption every abstraction shares.

Flow matching \cite{lipman2023flowmatching} trains a network $v_\theta(x, t, o)$ to transport a seed $\Xz \sim \mathcal{N}(0, I)$, of the same shape as the chunk, into data along the straight path $x_t = (1-t)\,\Xz + t\,\Xo$, $t \in [0, 1]$, by regressing its constant velocity:
\begin{equation}
  \mathcal{L}(\theta) = \mathbb{E}_{t,\,\Xz,\,(\Xo,\,o)}
  \big\lVert v_\theta(x_t, t, o) - (\Xo - \Xz) \big\rVert^2 .
  \label{eq:cfm}
\end{equation}
To execute a skill, we sample a seed and integrate $\dot{x} = v_\theta(x, t, o)$ from $t{=}0$ to $t{=}1$; the resulting chunk $\Xo = \Phi(\Xz; o)$ is executed open loop.
$\Phi$ is deterministic, so the seed and the observation determine the commanded chunk; its outcome also depends on contact physics, which the world model must absorb.
Within each batch, seeds are paired with chunks by minimum total distance, the minibatch optimal transport of \cite{tong2024otcfm}, so that nearby seeds train toward nearby chunks and the seed behind a chunk can be recovered and compressed, as abstraction (i) requires.

A single policy is trained on the pooled demonstrations of every skill, so the choice of skill is carried by the policy's inputs: the seed alone, or a code or label given with it.
A chunk spans a whole skill between two \emph{settled scenes}, in which the robot is stopped and the objects are at rest, as in quasi-static manipulation.
We call the configuration of the robot and the objects in a settled scene a \emph{state}, and let $o_k$ be its observation before skill $k$.
One step of the world model maps the observation and the action to the next observation,
\begin{equation}
  \hat f : (o_k, a_k) \mapsto o_{k+1},
  \label{eq:wm-step}
\end{equation}
where the action $a_k$ is a compact description of the policy's inputs, with whatever task knowledge the abstraction injects.

\subsection{Four action abstractions}
\label{sec:representations}

The policy's inputs split into the observation, which is measured, and the action, which the planner chooses: the seed and, when present, a code or label given to the policy as an additional input, $\Xo = \Phi(\Xz; o, \cdot)$.
Each abstraction is defined by that additional input and by the world-model action $a_k$ it induces; one equation below defines each, and Fig.~\ref{fig:teaser} summarizes them.
The action has a discrete part, a learned code $c$ or a symbolic label $y$, a continuous part $\hat a$, or both.
The continuous part is a low-dimensional latent over the seed, so that the planner searches a few numbers instead of a $T \times d$ seed.
It is fitted after the policy by a variational autoencoder, with an encoder $E_\phi$ on the executed chunk and a decoder $D_\phi$ to a seed:
\begin{equation}
\begin{aligned}
  \hat a &\sim E_\phi(\Xo, o, \cdot), \\
  \mathcal{L}(\phi) &= \big\lVert \Phi\big(D_\phi(\hat a; o, \cdot);\, o, \cdot\big) - \Xo \big\rVert^2 \\
  &\quad + \beta\, \mathrm{KL}\big(E_\phi \,\|\, \mathcal{N}(0, I)\big),
\end{aligned}
  \label{eq:vae}
\end{equation}
where $\cdot$ stands for the discrete input of the abstraction, if any.
The loss compares chunks, not seeds, because many seeds produce the same chunk.
The world model and the search procedure are shared by all four; the action and how its candidates are proposed, the action proposal, differ, and it is given for each.

To determine which skill an execution performed, we use a \emph{symbolic reference layer} external to every learned component: a vocabulary of skills with hand-written preconditions and effects.
A \emph{symbolic state} is the scene described in the vocabulary's terms, here which object rests where; the preconditions say in which symbolic states a skill applies, and the effects give the symbolic state after it.
Let $\sigma_k$ be the scene description in $o_k$, the part of the observation that describes the objects, here their poses.
From it the layer lists the skills available there, $\Asym(\sigma_k)$; from two consecutive scenes it reports which skill was performed, $\Cclass(\sigma_k, \sigma_{k+1}) \in \Asym \cup \{\text{none}\}$.
No learned component reads it; it is used for evaluation and by the action proposal of abstraction (iv).

\smallskip\noindent\textbf{(i) Pure seed.}
\label{sec:x0-action}
The policy receives only the seed, $\Phi(\Xz; o)$, so the seed carries both which skill is performed and how; this is the abstraction forced by unlabeled demonstrations, and the most general one.
The action is a compressed seed: a latent fitted by Eq.~\eqref{eq:vae}, decoded into a seed at execution,
\begin{equation}
\begin{aligned}
  a_k &= (c_k, \hat a_k), \\
  c_k &= \mathrm{VQ}\big(E_\phi(\Xo_k, o_k)\big), \qquad \hat a_k \sim E_\phi(\Xo_k, o_k), \\
  \Xz_k &= D_\phi(c_k, \hat a_k;\, o_k),
\end{aligned}
  \label{eq:a-seed}
\end{equation}
with a discrete code $c$ from vector quantization (VQ) \cite{vandenoord2017vqvae} for the skill and a Gaussian $\hat a$ for the residual.
The discrete part is needed because a plain Gaussian prior concentrates on the most common skill and rarely reaches the others; the latent widths are set by a reconstruction sweep in the experiments.
Candidates are proposed without task knowledge: the planner enumerates $c$ and searches $\hat a$.

\smallskip\noindent\textbf{(ii) Learned code.}
\label{sec:code-action}
Abstraction (i) extracts its action from a policy trained without any notion of a skill.
Here a \emph{discrete learned code} is fitted \emph{before} the policy, from the demonstrations alone.
A skill encoder $E_\psi$ maps the observations at the start and end of a demonstration to a code, and a skill decoder $D_\psi$ learns to predict the scene description after the skill from the start observation and the code:
\begin{equation}
\begin{aligned}
  c_k &= \mathrm{FSQ}\big(E_\psi(o_k, o_{k+1})\big), \qquad D_\psi(o_k, c_k) \approx \sigma_{k+1}, \\
  a_k &= \big(c_k,\, D_\psi(o_k, c_k)\big).
\end{aligned}
  \label{eq:a-code}
\end{equation}
The code is quantized by finite scalar quantization (FSQ) \cite{mentzer2023fsq}, which rounds each entry of a low-dimensional vector to fixed levels; $E_\psi$ is used only at training time.
The policy receives the code as its additional input, $\Phi(\Xz; o, c)$, and the seed is sampled fresh at execution, so $\Xz$ only varies how the skill is performed and enters the world model as execution noise.
The world-model action pairs the code with $D_\psi(o_k, c_k)$: a code describes an effect relative to the state it was learned in, and the prediction makes that effect explicit for the world model.
Candidates are proposed by a code prior $p(c \mid o)$ fitted on the demonstrations; at later steps of the search, where no observation exists yet, the prior reads $D_\psi(o, c)$ of the previous code, so the decoder plays the role of the symbolic effects, approximately.
The candidates then pass through a learned \emph{probabilistic precondition filter}, the counterpart of the preconditions: a classifier trained on the policy's own rollouts, with no symbolic label, that reads the world model's embedding of a state and predicts whether a code executed there produces $D_\psi(o, c)$.
Because it reads embeddings, it runs at every step of the search.
The assumption added is that a skill ends in a settled scene whose object poses are observed.

\smallskip\noindent\textbf{(iii) Object-centric code.}
\label{sec:ptr-action}
A code of (ii) describes an effect relative to the state it was learned in, so the same code applied elsewhere rarely reproduces the same skill.
Here the code refers to objects instead, so it means the same in every state; it is still learned from observations only:
\begin{equation}
  c_k = \big(i_k,\ j_k,\ g_k,\ r_k\big) = E_\psi(o_k, o_{k+1}), \qquad a_k = c_k,
  \label{eq:a-ptr}
\end{equation}
where $i$ points to the object that moves, $j$ to its destination, here another object or the table surface, $g$ is a coarse position on a fixed grid over that destination, and $r$ is one free factor, quantized as in (ii).
The two pointers are hard selections over the objects of the observation, trained with a straight-through gradient from the decoder loss of (ii).
The policy input, the action proposal, and the filter are as in (ii); the world-model action is the code alone, since it needs no prediction to fix its meaning.
The assumptions added are an observation that lists objects with stable identities, one moved object per skill, and the fixed grid.

\smallskip\noindent\textbf{(iv) Symbolic label.}
\label{sec:oracle-action}
Here each demonstration carries a symbolic label, and the policy receives it, $\Phi(\Xz; o, y)$, so $y$ selects what to do and the seed carries only how:
\begin{equation}
\begin{aligned}
  a_k &= (y_k, \hat a_k), \qquad y_k = (\text{kind}, \text{object}, \text{target}), \\
  \hat a_k &\sim E_\phi(\Xo_k, o_k, y_k),
\end{aligned}
  \label{eq:a-label}
\end{equation}
where the target is a location or another object, and $\hat a$ is a low-dimensional Gaussian latent over the residual, fitted by Eq.~\eqref{eq:vae} with $y$ as the discrete input.
Candidates are proposed by the preconditions and effects of the reference layer.
The assumptions added are a vocabulary of skills designed for the domain, a label for every demonstration, and those preconditions and effects.

\subsection{Necessary properties}
\label{sec:properties}

Two properties make an action usable, whichever part of it carries the skill: the seed in (i), the code or label in (ii) to (iv).
\textbf{Executability}: a sampled action completes some valid skill with high probability, measured as the \emph{valid-action mass}, the fraction of sampled actions that perform some valid skill, and every skill available in a state is reached by some action, measured as \emph{coverage} over a fixed number of samples.
\textbf{Decodability}: the skill an execution performs is predictable from the action and the observation.
For (ii) to (iv) decodability holds by construction, and executability reduces to the success rate when a given code or label is executed.
For (i) both must be established empirically: decodability is measured by a classifier on $(o, \Xz)$ against one trained on $o$ alone, so that only the information the seed adds is credited.
How well both survive the compression to $(c, \hat a)$ bounds its planning performance.

\section{World-Model Dynamics and Planning}
\label{sec:wm}

The robot plans in a receding-horizon loop: it observes the scene, imagines the outcome of candidate skill sequences, executes the first skill of the best sequence, and re-observes.
This requires a latent world model that predicts how an observation changes under an action (\S\ref{sec:lewm}), a planner that searches over action sequences in that model (\S\ref{sec:planning}), and a choice of what the observation contains (\S\ref{sec:state}).
The world model and the search procedure are shared by the four action abstractions of \S\ref{sec:flow-actions}; the action $a_k$ and how its candidates are proposed differ, which makes the comparison in \S\ref{sec:results} controlled.

\subsection{JEPA world model}
\label{sec:lewm}

The world model reads the same observation $o_k$ as the policy (\S\ref{sec:prelim}), the scene description $\sigma_k$ and the robot state, and predicts in a learned embedding space, following the joint-embedding predictive architecture (JEPA)~\cite{lecun2022jepa,assran2023ijepa}.
Nothing is reconstructed: the planner only compares imagined embeddings with the embedding of a goal, so the encoder only keeps what affects the next state.
The LeWM instantiation~\cite{maes2026leworldmodel} is used, which trains encoder and predictor end to end without a pretrained encoder or target network.
The chain evaluated at each planning step is
\begin{equation}
  o_k
  \;\xrightarrow{\ E\ }\;
  e_k
  \;\xrightarrow{\ P(\cdot,\, g(a_k))\ }\;
  \hat e_{k+1},
  \label{eq:pipeline}
\end{equation}
where $E$ is the encoder that maps the observation tokens to the embedding $e_k$, $g$ encodes the action $a_k$ into the same token format, $P$ is the predictor, and $\hat e_{k+1}$ is the predicted embedding of the next observation.
The $H$-step imagined rollout is
\begin{equation}
  \hat e_{k+i} \;=\; P\big(\hat e_{k+i-1},\ g(a_{k+i-1})\big),
  \qquad i = 1, \dots, H
  \label{eq:lewm-rollout}
\end{equation}
with $\hat e_k \equiv e_k$. Training minimizes the predictor's error against the encoder's own embedding of the observation that was actually reached,
\begin{equation}
  \mathcal{L}(E, P, g)
  = \mathbb{E}_{\mathcal{D}}
    \big\lVert P\big(E(o_k),\, g(a_k)\big) - E(o_{k+1}) \big\rVert^2
  \;+\; \lambda\, \mathcal{R}(E),
  \label{eq:lewm-loss}
\end{equation}
where $\mathcal{R}$ is the SIGReg regularizer~\cite{balestriero2025lejepa}, which prevents the embeddings from collapsing to a constant.

\subsection{Planning}
\label{sec:planning}

The goal is given as an observation $o^\star$ and passed through the same frozen encoder, $e^\star = E(o^\star)$, so that goal and predictions live in the same space.
Planning searches for the sequence of $H$ actions whose imagined state embeddings come closest to the goal,
\begin{equation}
  J\big(a_{k:k+H-1}\big) \;=\; \min_{h \le H} \big\lVert \hat e_{k+h} - e^\star \big\rVert^2 ,
  \quad
  a^\ast = \arg\min_{a} J(a),
  \label{eq:plan-cost}
\end{equation}
with $\hat e_{k+h}$ given by~\eqref{eq:lewm-rollout}.
The best state embedding along the rollout is scored, not only the last, so that a sequence may reach the goal early and hold.
Ties are broken in favor of the sequence that reaches its best state embedding in the fewest actions. 
The minimum is scaled by up to $1\%$ in proportion to the step at which it is reached, so a sequence whose minimum is more than $1\%$ lower wins regardless.

How~\eqref{eq:plan-cost} is optimized depends on which parts the action has (Fig.~\ref{fig:teaser}).
For abstractions (ii)--(iv), whose action contains a discrete code or label, a beam search enumerates discrete sequences up to horizon $H$ and keeps the most promising ones.

For abstractions (i) and (iv), whose action contains the continuous latent $\hat a$, the cross-entropy method (CEM) samples the continuous parts, scores the resulting rollouts via~\eqref{eq:lewm-rollout}, and refits its sampling distribution toward the lowest-cost samples, repeating for a few iterations.
In abstraction (iv) the two are nested: CEM runs within each sequence kept by the beam \cite{gimenez2026cross}.
Along the search, the world model imagines embeddings only, but an action proposal needs a state: abstraction (iv) carries it with the symbolic effects, (ii) and (iii) with the end state their skill decoder predicts, and (i) needs none because it enumerates.
These states serve the proposal only; every sequence is scored by the world model's imagined embedding, and the filter of (ii) and (iii) reads that embedding too.
Only the first action $a_k^\ast$ of the best sequence is executed: its continuous part, if any, is decoded into a seed, otherwise a seed is sampled, and the frozen flow policy runs it open loop, after which the robot observes $o_{k+1}$ and plans again.

\subsection{Scene representation}
\label{sec:state}

The scene tokens $\sigma_k$ are the object poses.
We choose poses because they are the lowest-dimensional description of the scene that still determines the task: the world model stays small enough to train on the available transitions, its predictions can be checked against the true poses, and any difference between the four action abstractions is attributable to the action rather than to the state.
The world model above does not depend on this choice; the scene tokens can also be image embeddings.
To deploy from images, a perception model $\Pi$ regresses the object poses from fixed camera views, $\hat\sigma_k = \Pi(I_k)$, trained on the demonstration images with their recorded poses and frozen before the world model is trained.
It is the only component that touches pixels; the world model, the planner, the goal $o^\star$, and the flow policy's observation (\S\ref{sec:prelim}) can all receive $\hat\sigma_k$ from it.
The results use the true poses, the upper bound, and report the perceived poses $\hat\sigma_k$ as the deployable case, so that the cost of perception is measured separately from the cost of the action abstraction.

\section{Results}
\label{sec:results}

\subsection{Experimental setup}
\label{sec:results-setup}

\begin{figure}[t]
  \centering
  \includegraphics[width=\linewidth]{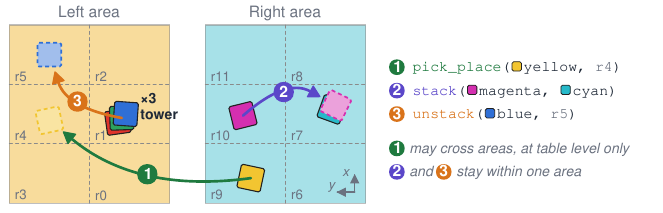}
  \caption{\textbf{The block scene}, top view, with one instance of each skill; dashed blocks are goal poses.}
  \label{fig:setup}
\end{figure}

\noindent\textbf{Scene.}
The scene is a simulated tabletop (MuJoCo) with a Franka Panda arm and six colored blocks (4\,cm cubes) on a table split into two areas of six regions each (cf.\ Fig.~\ref{fig:setup}).
The skills are $\text{pick\_place}(b, r)$, $\text{stack}(b, b')$, $\text{unstack}(b, r)$, and $\text{noop}(\cdot)$, which allows the robot to stay still; the symbolic reference layer lists their valid instances per state.
A block can be moved only when no block rests on it, stack and unstack stay within one area, and a tower holds at most four blocks. Finally, a target region must be free.

\smallskip\noindent\textbf{Dataset and flow policies.}
A scripted expert provides $120$k demonstration chunks of $T{=}256$ setpoints with $d{=}10$ values (position, 6-D rotation \cite{zhou2019rot6d}, gripper), branched so that $16$ different valid skills are demonstrated from each of $7\,500$ start states.
The four flow policies share this data and the same network, a transformer over the chunk with the observation injected by cross-attention; they differ only in the additional input (none, the learned code, the object-centric code, the label).
The pure-seed latent has a $256$-code discrete part and a $4$-d Gaussian part; the learned code has $625$ values; the object-centric code has $1512$; the labeled residual is $2$-d.
World-model dataset transitions come from rolling out each flow policy in episodes of $25$ skills: $51$k for the label, $64$k for the pure seed, and $128$k for each learned code; one world model is trained per abstraction with the same recipe.

\smallskip\noindent\textbf{Evaluation tasks and protocol.}
Tasks come from a fixed problem set: for every difficulty $k = 1, \dots, 14$, $100$ start states with a goal arrangement whose true minimum number of skills is exactly $k$, computed by exact search over the symbolic reference layer; $k{=}15$ is unreachable with six blocks.
Every abstraction is evaluated on the same tasks, so all curves are paired.
A task allows $k{+}5$ turns; the lookahead at each turn is the number of turns left, capped at $15$.
The beam search keeps $32$ sequences per level; CEM uses $64$ samples for $5$ iterations on the labeled residual and $128$ samples for $10$ iterations on the pure-seed latent.
The metric is symbolic success: the final symbolic state equals that of the goal arrangement.
We report Wilson $95\%$ intervals and paired McNemar tests between curves.
Unless stated otherwise, the world model and the policy receive the true poses, the upper bound; the perceived pass feeds both from the perception model on three fixed cameras (front, side, wrist), chosen so that no block is hidden from all of them, with proprioception from the robot.

\smallskip\noindent\textbf{Control-rate baseline.}
The low-level baseline is the world model of \S\ref{sec:wm} over control-rate actions instead of skills.
Its observation is the block poses extended with the arm state (end-effector position, $6$-D rotation, and gripper width; $64$-d in total), and its action is one absolute end-effector setpoint, $a_t = (p_t, R_t, g_t) \in \mathbb{R}^{10}$ (position, $6$-D rotation, gripper command), held for $7$ control steps ($0.35$\,s at $20$\,Hz), so one skill spans $38$ such steps.
It is trained on rollouts of the same frozen labeled flow policy, recorded at control rate with the commanded setpoints, and evaluated on the same task set with the same budget in physical time: a task at difficulty $k$ allows the $k+5$ turns of the skill-level planner, i.e. $266\,(k+5)$ control steps or $38\,(k+5)$ planner steps.

\begin{figure}[t]
  \centering
  \includegraphics[width=\linewidth]{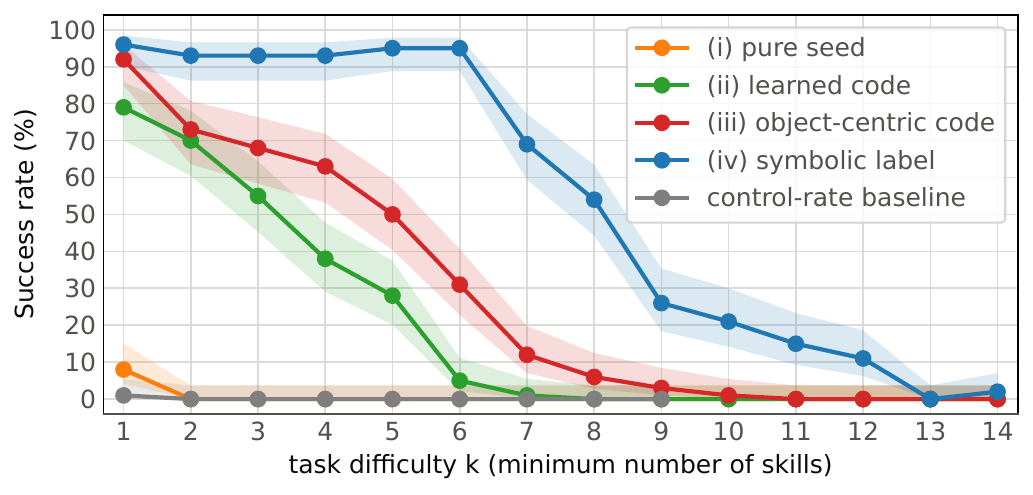} %
  \caption{\textbf{Planning success versus task difficulty $k$} ($100$ paired tasks per $k$, $k{+}5$ turns, Wilson $95\%$ intervals). Success grows with the injected knowledge; every abstraction falls from $k \approx 7$.}
  \label{fig:success_vs_k}
\end{figure}

\subsection{Planning success versus task difficulty}
\label{sec:results-planning}

Fig.~\ref{fig:success_vs_k} shows success against task difficulty, and the ordering follows the injected knowledge.
The pure seed solves $8\%$ of the tasks at $k{=}1$ and none beyond.
The learned code solves $79\%$ at $k{=}1$ but falls to $5\%$ by $k{=}6$.
The object-centric code matches the label at $k{=}1$ ($92\%$) and keeps one half to three quarters of the label's success from $k{=}2$ to $5$.
The symbolic label plans reliably up to six skills, $93$ to $96\%$ for $k \le 6$, and falls from $k{=}7$.
Paired tests confirm the ordering: the label is above every other abstraction for $k = 2$ to $12$ (McNemar, $p < 0.001$) and level with the object-centric code at $k{=}1$ ($p = 0.22$); beyond $k{=}12$ all four are near zero.

\smallskip\noindent\textbf{Control-rate baseline.}
Planning one control-rate action at a time, the baseline struggles to complete one skill.
We report a success rate of $1\%$ at $k{=}1$ and $0\%$ at $k > 1$ ($100$ tasks per $k$), and neither a longer horizon nor a larger planning budget changes this.
Success depends on the distance between the start and goal positions of the moved block: it is $9\%$ for the shortest moves ($5.5$ to $10$\,cm), and beyond this distance it drops to $0\%$.
We validate this result with off-the-shelf LeWM and DINO-WM \cite{maes_lld2026swm} implementations, trained and evaluated on single-camera pixel observations of the OGBench \texttt{cube-single} task.
The pick-and-place task spans start-to-goal distances from $7.5$ to $43$\,cm; both models reach a success rate of at most $12\%$ at the shortest distance and $0\%$ beyond.

\begin{figure}[t]
  \centering
  \includegraphics[width=\linewidth]{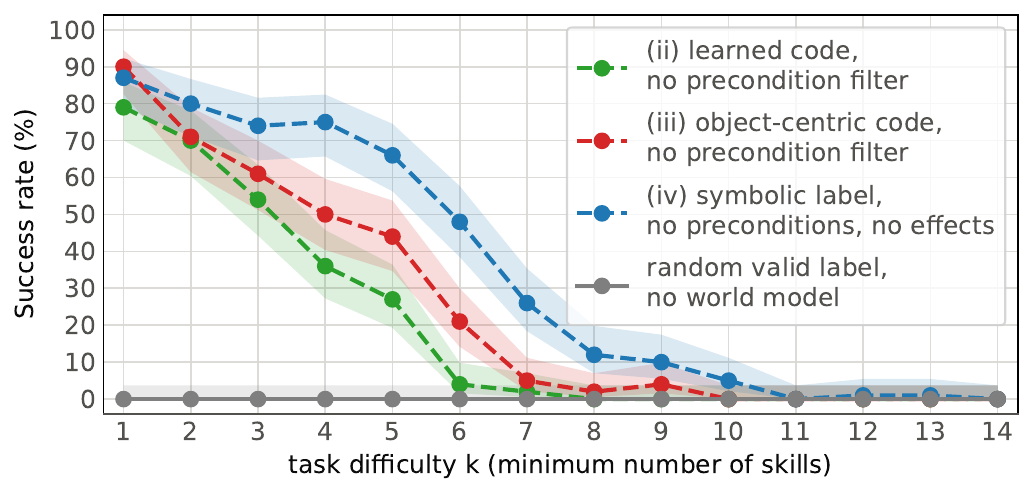}
    \caption{\textbf{Validity of the proposed candidates.} The label and the two codes without the precondition knowledge in their proposal, dashed, colors as in Fig.~\ref{fig:success_vs_k}; the label also loses its effects. Gray: a random valid label at each turn, without world model. All fall below Fig.~\ref{fig:success_vs_k}, the label the most.}
  \label{fig:validity}
\end{figure}

\subsection{Quality of the action abstractions}
\label{sec:results-actions}

Table~\ref{tab:action-quality} measures the two properties of \S\ref{sec:properties} per abstraction, and how well the world model tells actions apart.

The pure seed has both properties and loses part of them in compression.
A fresh seed performs a valid skill in $57\%$ of the executions and reaches every skill available in a state (coverage $100\%$ over $2048$ seeds); a classifier on $(o, \Xz)$ recovers the skill performed in $33\%$ of the cases among $174$ classes, against $8\%$ from $o$ alone.
Compression to $(c, \hat a)$ halves the valid mass to $30\%$, and the prior concentrates on the most common skill: stack and unstack are $3.7\%$ of its samples, against $15\%$ of fresh seeds.

Given a code or a label, the policy performs its skill in $90\%$ (learned), $92\%$ (object-centric), and $98\%$ (label) of the executions on states not seen in training; a fresh seed changes the outcome in $5\%$ of them, which the world model absorbs as execution noise.
The precondition filters, trained on the world-model rollouts and reading its embedding, separate the codes that can be executed from those that cannot with an area under the ROC curve (AUC) of 0.73 and 0.78; 
the search considers the codes whose probability is at least $0.5$, never fewer than $4$, and always $\text{noop}$.
Fixing the label's residual to zero and skipping CEM changes planning success by at most $5$ points at every $k$: the residual matters only for precise placement on a tower.

The world model separates the actions: on transitions not seen in training it ranks an action with the executed effect first in $98\%$ (label), $60\%$ (learned code), and $59\%$ (object-centric code) of the cases, the ordering seen in planning.
\begin{table}[t]
  \centering
\caption{Action quality per abstraction (\S\ref{sec:properties}), in percent. Two values: fresh seed / sample from the latent prior.}
  \label{tab:action-quality}
    \footnotesize
  \setlength{\tabcolsep}{5pt}
  \resizebox{\columnwidth}{!}{%
  \begin{tabular}{@{}lcccc@{}}
    \toprule
    Abstraction & Executable & Decodable & Precondition model & Ranked first \\
    \midrule
    (i) pure seed           & 57 / 30 & 33 & none & --- \\
    (ii) learned code        & 90 & by construction & filter (AUC 0.73) & 60 \\
    (iii) object-centric code & 92 & by construction & filter (AUC 0.78) & 59 \\
    (iv) symbolic label      & 98 / 95 & by construction & hand-written & 98 \\
    \bottomrule
  \end{tabular}%
  }
\end{table}

\subsection{What limits the horizon}
\label{sec:results-limits}

Every action proposal in the planner offers candidate actions and the world model \emph{scores} them.
The four proposals differ in what they know about which actions are applicable in a state, which we call \emph{precondition knowledge}: the label's proposal has the hand-written preconditions and effects of the vocabulary, so it offers only valid labels and knows the symbolic state after each; the codes' proposal has a classifier, trained on the same rollouts as the world model, that predicts from the embedding whether a code will do what its decoder says, and carries the state forward with the decoder's predicted end state; the pure seed has neither.

\smallskip\noindent\textbf{Valid candidates: is the label's advantage the label, or its planner's preconditions?}
We remove the precondition knowledge from the proposal and keep the action, the world model, and the search unchanged (Fig.~\ref{fig:validity}).
For the label, the proposal then offers every label whose block and target exist, at every step of the search, and no longer carries the symbolic state forward with the symbolic effects, which are defined for valid labels only.
For the object-centric code, the filter is removed and the skill decoder's predicted states remain.
Compared to Fig.~\ref{fig:success_vs_k}, the label loses the most, the object-centric code less, and the learned code almost nothing, so the two curves without preconditions nearly meet where the full curves are far apart.
Most of the label's advantage at short horizons is therefore its proposal knowing which labels apply, not the label as an action: preconditions and effects are worth 10 to 47 points to the label, the learned filter 2 to 13 to the object-centric code.

What remains after the ablation is the world model's scoring.
With valid labels but no world model, a random valid label at each turn, success is $0\%$ at every $k$: scoring is what turns valid candidates into plans.
Scoring is better when the action names its effect: on transitions not seen in training, the world model ranks the executed label first among the valid ones in $98\%$ of the cases, against about $60\%$ for either code (Table~\ref{tab:action-quality}).
The gap that remains between the codes and the label is therefore ranking, not validity.

\smallskip\noindent\textbf{Informative candidates: why does every abstraction fail for longer horizons?}
Success decreases at longer horizons, but accurate world-model predictions alone do not ensure that the planner finds a useful action sequence.
For a $k{=}14$ task, rolling out an optimal plan yields an MSE cost of $7$ to $20$ under Eq.~\eqref{eq:plan-cost}, against about $300$ for random action sequences.
The model can therefore identify promising plans when they are proposed.
Finding them is harder: beam search retains partial plans with the lowest predicted distance to the goal, although necessary intermediate moves may leave this distance unchanged or temporarily increase it.
Such plans can be pruned before their benefit becomes apparent.

\begin{figure}[t]
  \centering
  \includegraphics[width=0.95\linewidth]{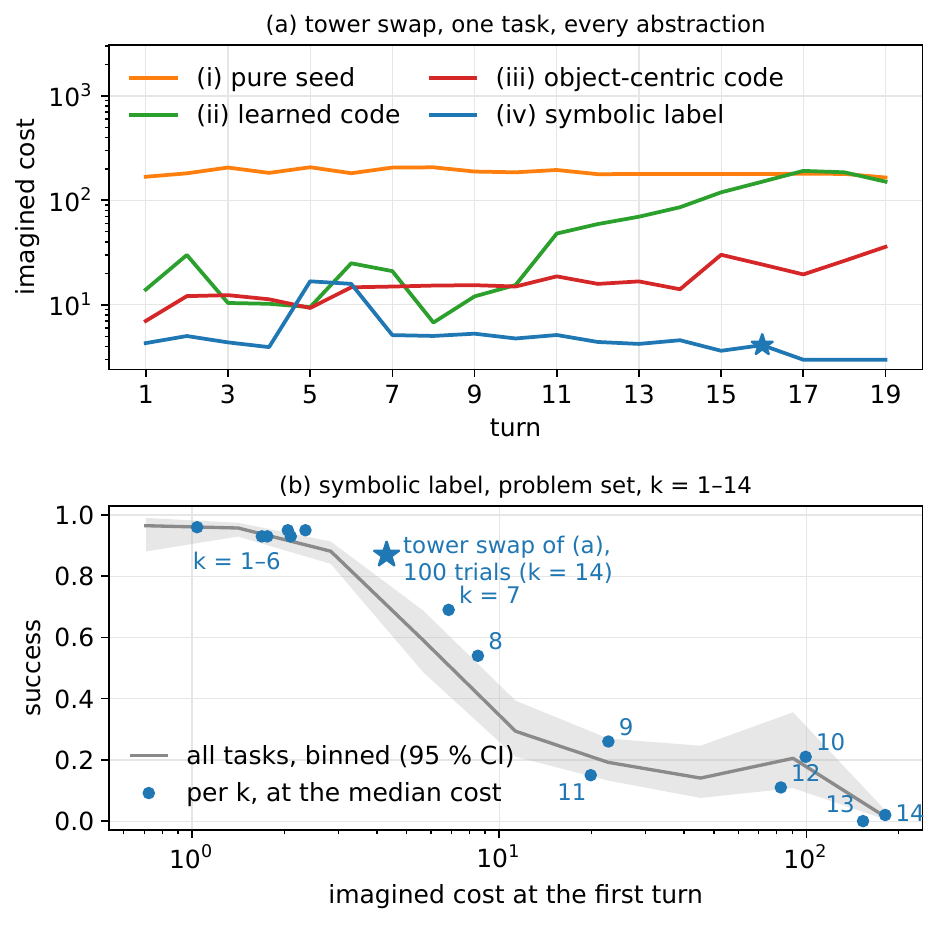}
  \caption{\textbf{Imagined cost of the plan the beam returns.}
  (a)~The tower swap of Fig.~\ref{fig:banner}, one task: the cost of the chosen plan per turn, per abstraction. The star marks the turn at which the label reaches the goal.
  (b)~Symbolic label on the problem set, $k{+}5$ turns: success against the imagined cost of the plan returned at the first turn, pooled over all $1400$ tasks in log-spaced cost bins (grey, $95\%$ Wilson interval) and per $k$ at the median cost (dots). A low first-turn cost means the beam found a sequence the model believes reaches the goal, and success follows that cost rather than $k$.}
  \label{fig:showcase-cost}
\end{figure}

Fig.~\ref{fig:showcase-cost} relates the cost of the plan found at the first planning step to subsequent task success.
Even at $k{=}14$, the symbolic label solves the tower-swap example in $87\%$ of trials, with a median initial plan cost of $4.3$.
Here, staging moves begin to reduce the predicted distance early enough for search to retain a successful sequence.
This shows that the longest tasks remain solvable when search finds an informative plan at the outset.
A low predicted cost is not sufficient for every abstraction, however: the learned codes can propose promising plans but fail during execution.
Increasing the beam width from $32$ to $256$ does not improve success, suggesting that more candidates alone does not resolve the limitation of ranking partial plans by goal distance.

\begin{figure*}[t]
  \centering
  \includegraphics[width=\textwidth]{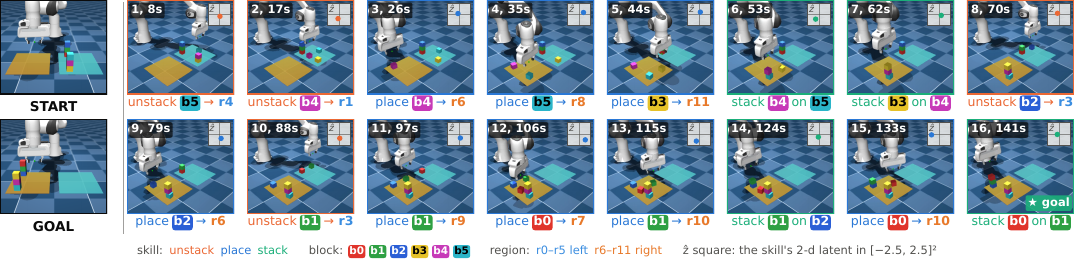}
  \caption{\textbf{The skill-level world model with the symbolic label solving the $k{=}14$ tower swap.} Left: the start and goal observations (front camera), the planner's only task specification. Right: one frame per executed skill, numbered by turn as in Fig.~\ref{fig:showcase-cost}, with the skill performed and, in the corner, the residual $\hat{a}$ of that skill. The planner reaches the goal after $16$ skills: it corrects two placements (turns 13 and 15) on the way. In this layout the staging moves of the optimal plan lower the imagined cost from the fourth action on, so the beam keeps them.}
  \label{fig:banner}
\end{figure*}

\subsection{Cost of perception}
\label{sec:results-perception}

With perceived poses in place of the true poses, success changes little.
Averaged over $k = 1$ to $6$, the label loses $1$ point, the learned code $1$, and the object-centric code $6$, with its only significant loss at $k{=}1$ ($92$ to $80\%$); beyond $k{=}6$ all differences are within noise, and the pure seed stays near zero.
The perception model's error on settled scenes is $3.0$\,mm in position (median $2.3$\,mm) and $2.5^\circ$ in rotation, symmetry-reduced, and the perceived symbolic state equals the true one in $90\%$ of the scenes.

\section{Conclusion}
\label{sec:conclusion}

We presented a skill-level world model in which one transition is one complete execution of a flow policy.
Four action abstractions built on the same policy were compared with the same world model and search procedure.
The symbolic label plans best, and an object-centric code learned from observations keeps much of its success at short horizons without any label.
The pure seed carries the skill, but in a form that neither its compression nor the world model recovers, and it does not plan.
Ablations attribute a substantial part of the label's advantage to the precondition knowledge in its action proposal, not to the label as an action.
Every abstraction loses reliability as difficulty grows, although the symbolic label still solves the tower swap at the highest evaluated difficulty when the layout lets its search see progress.
The search, not the world model, sets the horizon: the world model assigns the optimal plan a low cost, but a beam ranked by distance to the goal discards the intermediate moves that do not reduce it.
These results motivate action proposals and search objectives that recognize necessary intermediate moves, alongside more reliable skill execution.
Extending the method beyond segmented skills and settled scenes remains future work.

\bibliographystyle{IEEEtran}
\bibliography{references}

\end{document}